\documentclass{article}

\usepackage[square,numbers]{natbib}

\usepackage[preprint]{neurips_2021}

\usepackage[utf8]{inputenc} 
\usepackage[T1]{fontenc}    
\usepackage{hyperref}       
\usepackage{url}            
\usepackage{booktabs}       
\usepackage{amsfonts}       
\usepackage{nicefrac}       
\usepackage{microtype}      
\usepackage{xcolor}         
\usepackage{graphicx}
\usepackage{multirow}
\usepackage{amsmath,amssymb}
\DeclareMathOperator{\E}{\mathbb{E}}
\DeclareMathOperator*{\median}{median}
\usepackage{subcaption}

\title{Negotiating Networks in Oligopoly Markets for Price-Sensitive Products}

\author{%
  Naman Shukla \\
  Deepair Solutions\\
  \texttt{naman@deepair.io} \\
  \texttt{} \\
  \And
  Kartik Yellepeddi \\
  Deepair Solutions\\
  \texttt{kartik@deepair.io} \\
}

\begin{document}

\maketitle

\begin{abstract}
We present a novel framework to learn functions that estimate decisions of sellers and buyers simultaneously in a oligopoly market for a price-sensitive product. In this setting, the aim of the seller network is to come up with a price for a given context such that the expected revenue is maximised by considering the buyer's satisfaction as well. On the other hand, the aim of the buyer network is to assign probability of purchase to the offered price to mimic the real world buyers' responses while also showing price sensitivity through its action. In other words, rejecting the unnecessarily high priced products. Similar to generative adversarial networks, this framework corresponds to a minimax two-player game. In our experiments with simulated and real-world transaction data, we compared our framework with the baseline model and demonstrated its potential through proposed evaluation metrics.
\end{abstract}

\section{Introduction}

Markets dominated by a small number of sellers, none of which can keep others from having significant influence, are known as oligopoly markets \cite{shubik1959strategy}. In such markets, the prices of the offered services and products are usually set by the sellers, and buyers can only influence overall demand. Due to this, demand plays an important role in price determination and vice-versa. Revenue management has been instrumental for price determination in markets like hospitality, airlines, financial services, medical services, and the telecommunications industry \cite{talluri2004theory, boyd2003revenue}. Generally, prices are determined using demand forecasting \cite{daganzo2014multinomial} and estimating utility functions \cite{kim2002modeling} within these systems. Recently, it has been demonstrated that demand forecasting models based on deep neural networks outperformed traditional statistical techniques \cite{law2000back}. Also, discrete choice models trained using machine learning methods have proven to perform significantly better than conventional alternatives \cite{west1997comparative, imai2009bayesian}. Despite their respective advancements, there exist no framework which could leverage the potential of real-world market interactions between sellers and buyers. Furthermore, most of the pricing models are sequentially coupled with demand forecasting techniques which makes implementing real-world market negotiations even challenging. 

In this work, we present a novel framework to simulate oligopoly marketplace interactions. Using this setting, one can (a) learn two disentangled estimator functions representing seller and buyer actions respectively, (b) determine better optimal contextual prices and their corresponding demand, (c) explore new stable and unstable equilibrium prices, and (d) perform choice modeling with respect to context and price. 

\section{Related work}

  Topics in dynamic pricing of homogeneous products have been extensively studied \cite{den2015dynamic}. In fact, dynamic pricing has been a catalyst for innovation in various transport and service industries. Ride-hailing platforms have used surge pricing to match demand and supply, and to avoid the "wild-goose chase" problem \cite{castillo2017surge}. Customer characteristics and the seller's ability to price discriminate are also shown to significantly influence overall revenue and profit \cite{varian1989price}. It has been demonstrated that dynamic pricing using buyers' context can achieve higher conversion rates as well as revenue per offer by the sellers \cite{ye2018customized, shukla_dynamic_2019}. As far as buyer behaviour modeling is concerned, multiple theoretical models \cite{KahnemanTversky,GabaixLaibson,ShulmanGeng} based on risk perception, knowledge levels and bounded rationality; and discrete choice models \cite{BenAkiva} are typically used. Recently, it has been shown
  that buyer models based on pointer networks, first introduced by \citet{vinyals2015pointer}, offers a promising alternative for choice modeling \cite{mottini2017deep}.

Deep neural networks have demonstrated tremendous success in decision making across mass-impact domains,  such as finance \cite{nelson_stock_example}, pricing \cite{chiarazzo2014neural, ye2018customized} and policy-making \cite{hochtl2016big}. Among others, generative adversarial networks \cite{goodfellow2014generative} have emerged as powerful frameworks across multiple fields of application. For instance, some of the state-of-the-art models in computer vision for image segmentation \cite{chen2018semantic}, image-to-image translation \cite{isola2017image, zhu2017unpaired, liu2017unsupervised}, data augmentation \cite{bousmalis2017unsupervised} and classification tasks \cite{aggarwal2021generative} are based on generative adversarial nets. We derive inspiration from the generative adversarial framework, where two models compete with each other and thereby learn from each other. Similarly, models negotiate with each other (as they do in real market) and find the optimal price using a negotiating networks framework. 

\section{NegoNets: Negotiating Networks}

Consider the general setting of an oligopoly marketplace where buyers and sellers interact. 
A set of $n$ such interactions is represented by
\{($x_i, p_i, y_i$)\}, $i = 1, \dots, N$. Here, $\{x_i\}_{i=1}^N$ where $x_i \in \mathbb{R}^\mathcal{D}$ represent contextual information; $\{p_i\}_{i=1}^N$ where $p_i \in [0, 1]$ represents scaled offered price ($0$ and $1$ for minimum and maximum allowed prices respectively); and finally, the buyer's decision as probability of purchase is represented by $\{y_i\}_{i=1}^N$, where $y_i \in \{0, 1\}$. The objective of the method presented below is to determine two estimator functions $f_s$ and $f_b$ which approximate the seller's and buyer's decisions respectively. The seller suggests a price $f_s(x)$ for a context $x$ and the buyer takes this context along with the suggested price to assign a probability distribution , $f_b(x, f_s(x))$, over purchase or no-purchase options.

\subsection{Adversarial loss}

 To incorporate the negotiation between two networks, we apply adversarial loss to both estimator functions. For function $f_s$ : $\mathbb{R}^\mathcal{D} \rightarrow [0, 1]$ and $f_b$ : $\mathbb{R}^{\mathcal{D}+1} \rightarrow [0,1]$, we express the objective as:

\begin{equation}
\mathcal{L}_\text{ADV}(f_s, f_b) = \E\left[f_s(x) \cdot f_b(x, f_s(x)) - p \cdot y \right] + \E\left[ {\frac{\partial f_b(x, f_s(x))}{\partial f_s(x)}}^+ \right] 
\label{eq:adv_loss}
\end{equation}

where, $f_s$ tries to maximize the incremental expected revenue and $f_b$ aims to minimize the point wise loss \cite{gupta_how_2019}. The incremental expected revenue part of the loss function is responsible for making the seller push for higher prices in cases where the expectation of conversion is high. Whereas, point wise loss creates a shape constraint in the buyer network. Price sensitivity is induced using this shape constraint, which is responsible for penalizing negative correlated decisions from buyers network to recommended price. In other words, price sensitivity makes the buyer rational and stops the network from being compliant with higher prices.

\subsection{Independent loss}

Unlike adversarial loss, we introduce loss functions that are mutually independent. The independent loss function is expressed as:

\begin{equation}
\mathcal{L}_{\text{IND}}(f_s, f_b) =  \E \left[ \log(y, f_b(x, p)) \right] - \E \left[ \mathcal{B}(f_s(x), p, y) \right]
\label{eq:ind_loss}
\end{equation}

In equation (\ref{eq:ind_loss}), the first part is cross entropy loss for the buyer network and the second part is boundary loss, which is applied to the seller network. In equation (\ref{eq:ind_loss}), $\mathcal{B}$ represents the boundary function. This function is inspired from strategic model proposed by \citet{ye2018customized} and $\epsilon$-insensitive loss used in SVR \cite{smola2004tutorial}. The enhanced version of this function heuristically incorporates the monotonicity in the willingness to pay as mentioned in \citet{shukla_dynamic_2019}. The boundary function penalizes the networks for violating upper and lower bounds as follows:

\begin{equation}
\mathcal{B}(f_s, x, p, y) = \E \left[ (L(p, y) - f_s(x))^+ - (f_s(x), U(p, y))^+ \right]
\end{equation}

\begin{equation}
L(p, y) = y \cdot p + (1-y) \cdot c_1 \cdot p
\end{equation}
\begin{equation}
U(p, y) = (1-y)\cdot p + y \cdot c_2 \cdot p
\end{equation}

For all instances involving a purchase, the lower bound $L$ is the purchase price $p$. Otherwise, a lower price $c_1p$ is set to be the lower bound, where $c_1 \in (0,1)$. Similarly, the upper bound $U$ is $p$ for all the instances of no purchase. If there is a purchase, a price of $c_2p$ ($c_2>1$) is set as the upper bound. For the boundary loss to be non-zero, $c_2 > \frac{fs}{p}$ and $\frac{fs}{p} < c_1 < 1$. For $c_1 = c_2 = 1$, the lower bound and upper bound are equal and hence the trivial optimal price will be $p$. Therefore, $c_1$ and $c_2$ are hyperparameters that can be calibrated to change the gap between the lower and upper bounds. 

\subsection{Overall objective}

\begin{figure}
\centering
\includegraphics[scale=0.137]{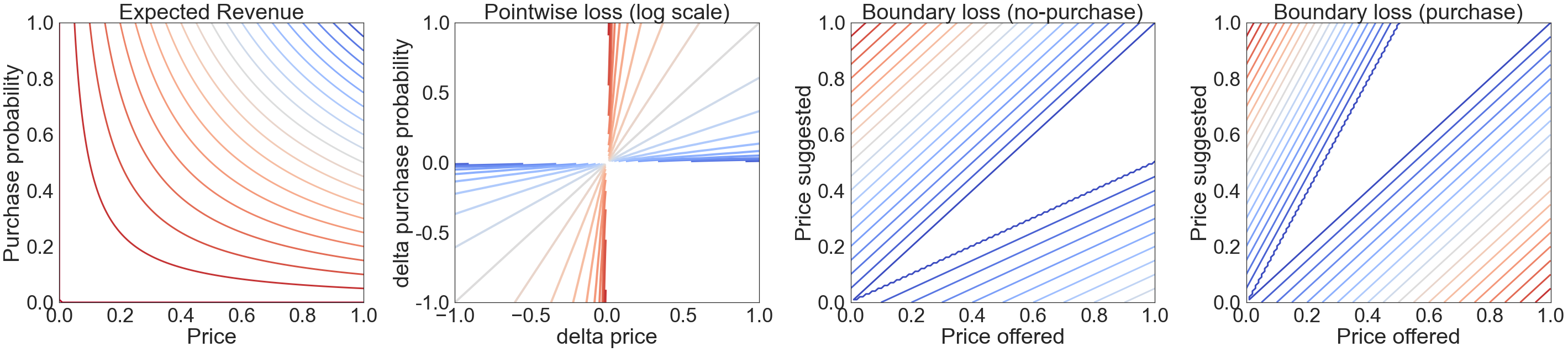}
\caption{Contour plots for the objective function. Here blue corresponds to low value and red corresponds to high value of the objective function. Starting from left, first two figure illustrates adversarial loss and last two illustrates boundary function from independent loss}
\label{fig:contours}
\end{figure}

The overall objective that we aim to solve is:
\begin{equation}
f_s^*, f_b^* = \text{arg} \min_{f_b} \max_{f_s} \mathcal{L}_{\text{ADV}} + \lambda \mathcal{L}_{\text{IND}} 
\label{eq:loss}
\end{equation}

where, $\lambda$ controls the relative importance of the two components. In practice, $\lambda$ can be calibrated to get desired amount of negotiation between the two network. For visual intuitions, the contour plots for each component of the loss function are presented in Fig \ref{fig:contours}. Expected revenue contour spans hyperbolic lines in price and probability of purchase space. Whereas, the point-wise loss, present in the first and the third quadrants of delta space (change in price and change in probability), represents price sensitivity. Finally, boundary function contour plots shows configurable optimal region in offered and suggested price space.


\section{Experiments}

In this section, we present empirical results for the proposed negotiating network framework as market place interaction. 
In addition to quantitative comparison, we perform post-hoc analysis of the method to gauge which configuration gives most realistic results. 
In the following experiments, we focused on one industrial application which could provide both simulated and real-world oligopoly market interactions. We aim to extend this analysis to other industries and markets to study the applicability in future research.

Demand forecasting is frequently used for price determination in price sensitive markets. Therefore, we use the buyer's baseline as estimated probability of an individual interaction converting to purchase, which, when aggregated over the population, is a proxy for estimated demand. Then, logistic regression is used for the given estimated demand to decide the selling price as the seller's baseline \cite{avlonitis2005pricing}. These baselines are then compared with proposed framework.

Across all experiments, we use 4-layer deep neural network for the buyer, the seller, and the baseline demand forecasting model. Adam \cite{adam_kingma} optimizer is used with an early stopping criterion. We tune hyperparameters such as optimizer and  learning rate, and keep them same for other networks to facilitate a fair comparison. Some method specific parameters like $c1$, $c2$ and $\lambda$ are set to be $0.5$, $2$ and $1$, respectively.

\subsection{Evaluation Metrics}

In this section, we define the metrics that are used for evaluating performance of the seller and the buyer models.  We use \texttt{F1} score to evaluate classification accuracy, and monotonicity score $\mathcal{M}_k$ \cite{gupta_how_2019, wilmott1995mathematics, gupta2021pender} to measure 
\emph{diminishing returns} (decreasing and convex shape) with respect to price increase for buyer model, as per utility theory.
Here, price $p$ is the target feature $k$. For seller models, we use the following custom metrics introduced by \citet{ye2018customized} and frequently used in applied revenue management \cite{ kolbeinsson2021galactic} to gauge the goodness of price:

 \textit{Price Decrease F1} (\texttt{PDF1}). This metric is inspired by the \texttt{F1} score used to evaluate the precision and recall tradeoff. \texttt{PDF1} therefore measures the tradeoff between \textit{Price Decrease Recall} (\texttt{PDR}) and \textit{Price Decrease Precision} (\texttt{PDP}) according to equation (\ref{eq:pdf1}). \texttt{PDR} measures how likely the suggested prices are lower than the actual offered prices for non-purchased interactions and \texttt{PDP} measures the percentage of non purchased interactions where the suggested price is lower than actual price.

\begin{equation}
\texttt{PDF1} = \frac{2 \cdot \texttt{PDR} \cdot \texttt{PDP}}{\texttt{PDR}+\texttt{PDP}}
\label{eq:pdf1}
\end{equation}

\textit{Price Increase F1} (\texttt{PIF1}). Similar to \texttt{PDF1}, this score uses \textit{Price Increase Recall} (\texttt{PIR}) and \textit{Price Increase Precision} (\texttt{PIP}) based on equation (\ref{eq:pif1}). \texttt{PIR} measures how likely the suggested prices are higher than actual offered prices for purchased interactions and \texttt{PIP} measures the percentage of purchased interactions where the suggested price is higher than actual price.

\begin{equation}
\texttt{PIF1} = \frac{2 \cdot \texttt{PIR} \cdot \texttt{PIP}}{\texttt{PIR}+\texttt{PIP}}
\label{eq:pif1}
\end{equation}

\textit{Regret Score} (\texttt{RS}). Defined by equation (\ref{eq:rs}), \texttt{RS} measures the central tendency of the missed opportunity to suggest the price greater or equal to actual offered price in all purchased interactions. 

\begin{equation}
\label{eq:rs}
\texttt{RS} = \median_{purchases}\Big(\max\big(0, 1-\frac{f_{s}}{p}\big)\Big)
\end{equation}

\subsection{Datasets}
 
  Ancillaries are optional products or services sold by businesses to complement their primary product \cite{bockelie2017incorporating}. In this work, we have utilized the following datasets from the airline industry for an ancillary market: (1) simulated interaction of flight seat (as ancillary) pricing using open-sourced flai simulator\footnote{https://github.com/deepair-io/flai}, and (2) real-world proprietary interactions for ancillaries from a large airline containing price variability.
  
\subsubsection{Simulator dataset}
 This dataset consists of simulated interactions for selecting flight seat after the ticket is purchased for a one-way flight. Customer arrivals are simulated using non-homogeneous poisson process \cite{lewis1979simulation}. Multinomial logit is used for customer choice model in the simulator \cite{hausman1984specification}. Offered prices for seat selection are randomly sampled from minimum to maximum allowed price for that flight. We have simulated 8 long-distance flights with high value of allowed prices for training, and 2 short-distance flights with low value of allowed prices.

\begin{table}[]
\centering
\begin{tabular}{c|cc|ccc}
\toprule
\multirow{2}{*}{Models} & \multicolumn{2}{c|}{Buyer}      & \multicolumn{3}{c}{Seller}                      \\
                        & \texttt{F1}             & $\mathcal{M}$              & \texttt{PDF1}           & \texttt{PIF1}           & \texttt{RS}             \\ \hline
NegoNets                 & \textbf{0.343 $\pm$}  0.088 & \textbf{1.000} & 0.173         & \textbf{1.000} & 1.089 \\
Baseline                & 0.161 $\pm$ 0.031         & 0.788 $\pm$ 0.100         & \textbf{0.400} & 0.190          & 0.989 
\end{tabular}
\caption{Model performance on flight seat pricing simulation test set}
\label{tab:sim}
\end{table}

According to Table \ref{tab:sim}, superior performance of buyer network trained using NegoNets as compared to baseline indicates that the network learns a generalised representation, which can account for change in test distribution i.e. from relatively less price sensitive interactions (higher allowed price value) to more price sensitive ones (lower allowed price value). While both NegoNets and baseline have similar \texttt{RS}, the seller model trained from NegoNets outperforms in \texttt{PIF1}. This implies that the seller network is able to increment prices for correctly identified purchased sessions. Whereas \texttt{PDF1} is slightly higher for baseline because estimated forecast is highly biased towards non purchased interactions, resulting in overall low suggested prices. 

\subsubsection{Real-world dataset}
Dynamic pricing of ancillary products and services is a recently adopted application in airline industry. We have used a proprietary real-world dataset for ancillary pricing from a large airline. This dataset contains transactions with offered prices varying from minimum value (normalized to 0) to maximum value (normalized to 1) for one of the ancillary services offered by the airline. Unlike simulations, there are 10 different origin-destination pairs with 680 unique flights in this dataset. We use 20\% of the dataset for testing, and perform a 80-20 split on the remaining data to create train and validation sets. It is worth emphasising that this dataset is inherently non-monotonic, i.e. as price increase the purchase ratio does not decrease. 

\begin{table}[hbtp!]
\centering
\begin{tabular}{c|cc|ccc}
    \toprule
    \multirow{2}{*}{Models} & \multicolumn{2}{c|}{Buyer}      & \multicolumn{3}{c}{Seller}                      \\
                            & \texttt{F1}             & $\mathcal{M}$             & \texttt{PDF1}           & \texttt{PIF1}           & \texttt{RS}             \\ \hline
    NegoNets                 & \textbf{0.682 $\pm$} 0.049 & 0.018 $\pm$  0.011        & 0.479         & \textbf{0.465} & \textbf{0.000} \\
    Baseline                & 0.626 $\pm$ 0.036        & \textbf{0.373 $\pm$} 0.003 & \textbf{0.511} & 0.353          & 1.126

\end{tabular}
\caption{Model performance on real-world airline ancillary test set}
\label{tab:real-world}
\end{table}

Interestingly, the seller network trained with NegoNets reached higher \texttt{F1} score with lower $\mathcal{M}$ as compared to baseline as shown in Table \ref{tab:real-world}. Since, the dataset is inherently non-monotonous and the shape constraints are not enforced explicitly but learned from the data, the higher \texttt{F1} score is justified. In other words, learning the \emph{diminishing returns} criteria could guide the network to attain higher local optima as compared to its exclusion. For seller models, both \texttt{PDF1} and \texttt{PIF1} scores are comparable. Yet, NegoNet seller is able to reach better \texttt{RS} as compared to its baseline. Reason being the baseline method optimizes price given the purchase probability, and the proposed method estimates the price distribution directly.   

\section{Discussion}
We show, for the first time, that oligopoly market interactions can be framed as negotiating networks. The experiments on airline ancillary simulation and on real-world dataset successfully demonstrates that learning disentangled networks (seller and buyer network) can achieve better individual performance as compared to sequential forecasting and optimization conventional technique. Besides this, we have shown that using negotiating networks framework, one can explore new ranges of potential prices along with their estimated demand to come up with novel policies and market equilibrium states. 

We have presented the results from our first attempt to model market interactions, and there is much room for improvement. In practice, the proposed framework is sensitive to model parameters. Since GAN-based networks are prone to mode collapse, such framework configuration becomes even critical. Hence, sustaining  equilibrium state in such settings is still an open question. In this framework, we have not included other confounding factors like cost of the product, inventory management, and competitor's influence on both sellers and buyers. Furthermore, contextual information conditioned on the network has not been explored yet. We plan to extend our empirical evaluation to other applied fields. We are specifically interested in further examining the attained pricing equilibrium with theoretical guarantees. Given that negotiations can be achieved based on price, it will help our understanding of whether other attributes similar to price make interaction more efficient or not.

\bibliographystyle{unsrtnat}
\bibliography{main}

\newpage
\appendix

\section{Appendix}

\subsection{Additional Information}

Here, we present some additional insights and analysis on datasets. We provide configuration files for reproducibility of the simulations.\footnote{https://drive.google.com/drive/folders/1lZeotPPjnwdoOp2H4qSuSXU0rh-2aUzC?usp=sharing}

\begin{figure}[ht] 
  \begin{subfigure}[b]{0.5\linewidth}
    \centering
    \includegraphics[scale=0.2]{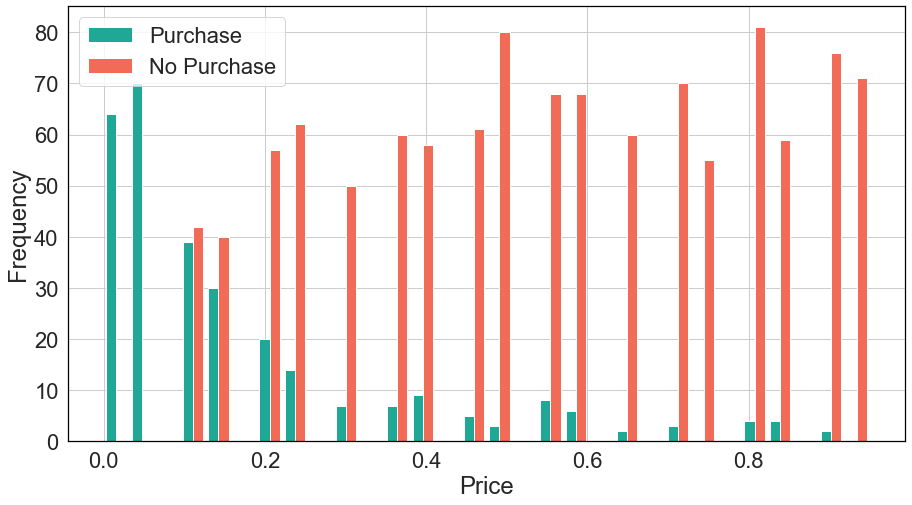} 
    \caption{Simulation train set} 
    \label{dist:a} 
    \vspace{4ex}
  \end{subfigure}
  \begin{subfigure}[b]{0.5\linewidth}
    \centering
    \includegraphics[scale=0.2]{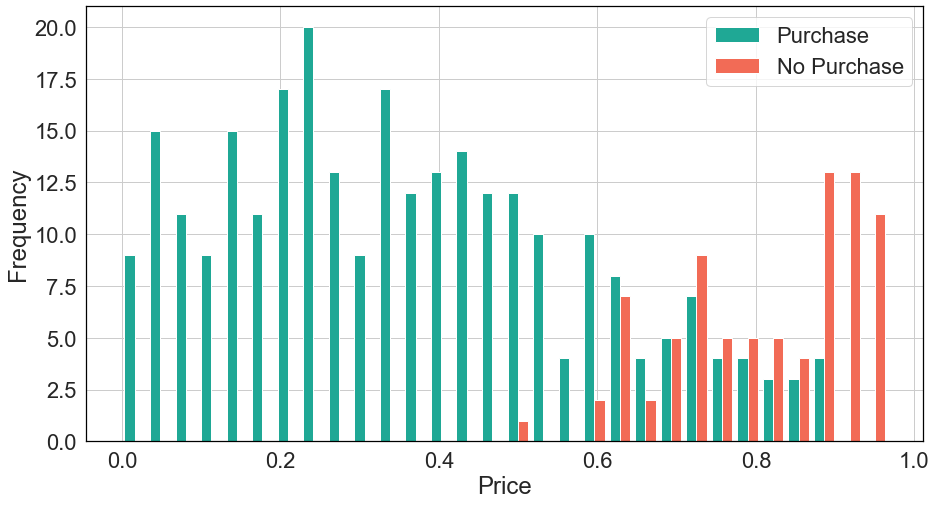} 
    \caption{Simulation test set} 
    \label{dist:b} 
    \vspace{4ex}
  \end{subfigure} 
  \begin{subfigure}[b]{0.5\linewidth}
    \centering
    \includegraphics[scale=0.2]{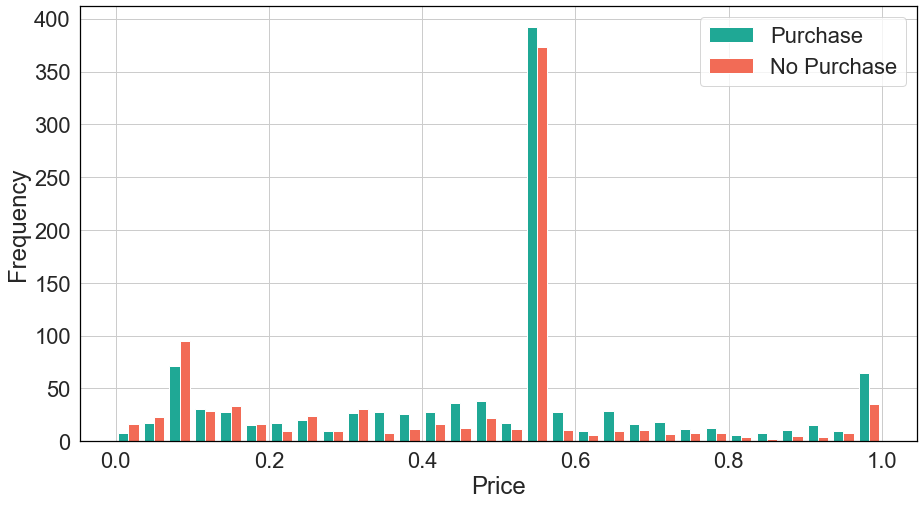} 
    \caption{Airline ancillary train set} 
    \label{dist:c} 
  \end{subfigure}
  \begin{subfigure}[b]{0.5\linewidth}
    \centering
    \includegraphics[scale=0.2]{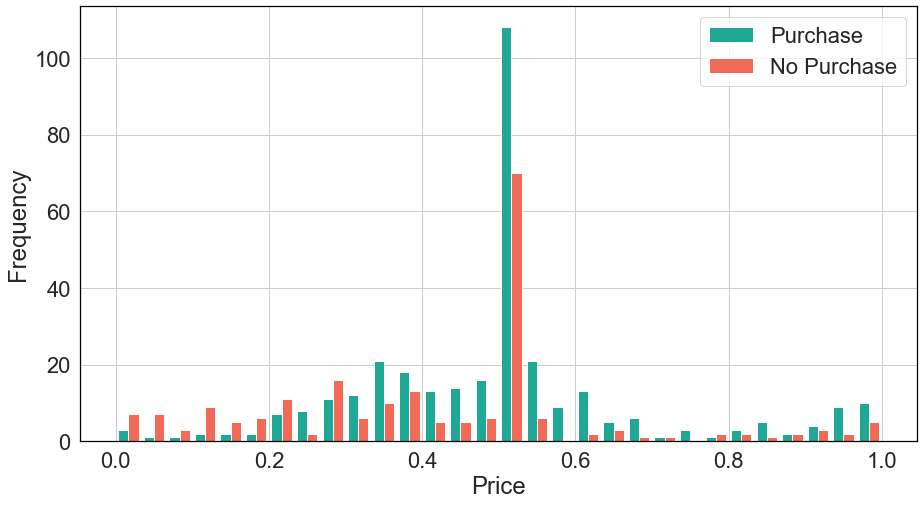} 
    \caption{Airline ancillary test set} 
    \label{dist:d} 
  \end{subfigure} 
  \caption{Histogram of offered prices (normalized) in the respective dataset}
  \label{dist} 
\end{figure}

Figure \ref{dist} shows the distribution of prices in the train and the test set for their respective datasets. It is worth mentioning that the minimum and maximum allowed price values are different for train and test set for simulation data represented in Figure \ref{dist:a} and \ref{dist:b}. Specifically, test set have a lower values of minimum and maximum allowed price than in train set. For airline ancillary dataset, the allowed price window is same for train and test set. Nevertheless, there is a significant difference in the frequency of the most common offered prices and the rest, as shown in Figure \ref{dist:c} and \ref{dist:d}.

\begin{figure}[ht]
    \centering
    \begin{subfigure}{0.5\textwidth}
        \centering
        \includegraphics[scale=0.2]{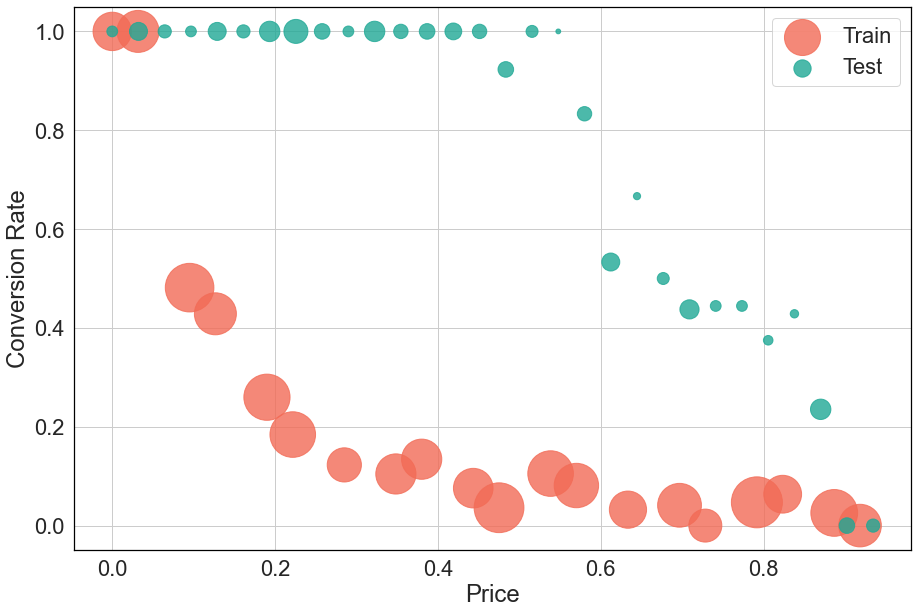}
        \caption{Simulation dataset}
        \label{fig:conv:a}
    \end{subfigure}%
    ~ 
    \begin{subfigure}{0.5\textwidth}
        \centering
        \includegraphics[scale=0.2]{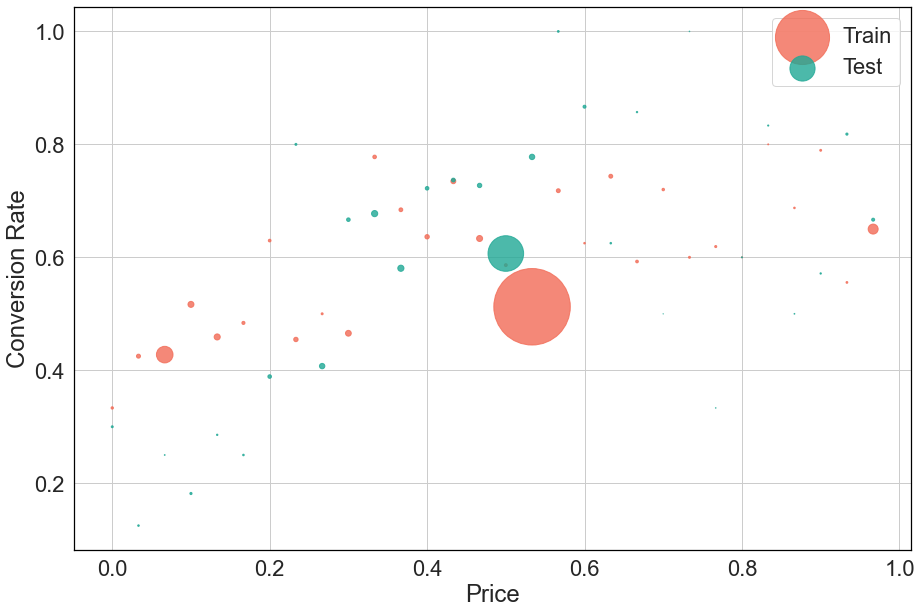}
        \caption{Airline ancillary dataset}
        \label{fig:conv:b}
    \end{subfigure}
    \caption{Conversion rates for offered price in the respective dataset. The radius of the scatter plot is proportional to frequency of event in that dataset.}
    \label{fig:conv}
\end{figure} 

The conversion rates shown in Figure \ref{fig:conv} validates the willingness to pay assumption in buyers i.e. if a buyer is willing to
buy a product for price $p$, they are willing to buy the same product at a price $p' < p$. Similarly, if a buyer is unwilling to buy a product at price $p$, they will be unwilling to buy at a price $p' > p$. Unlike simulated data in Figure \ref{fig:conv:a} which shows a decreasing trend, airline ancillary dataset shows non-decreasing trend in Figure \ref{fig:conv:b}. 
\subsection{Additional Observations}

\begin{figure}[ht] 
  \begin{subfigure}[b]{0.5\linewidth}
    \centering
    \includegraphics[scale=0.2]{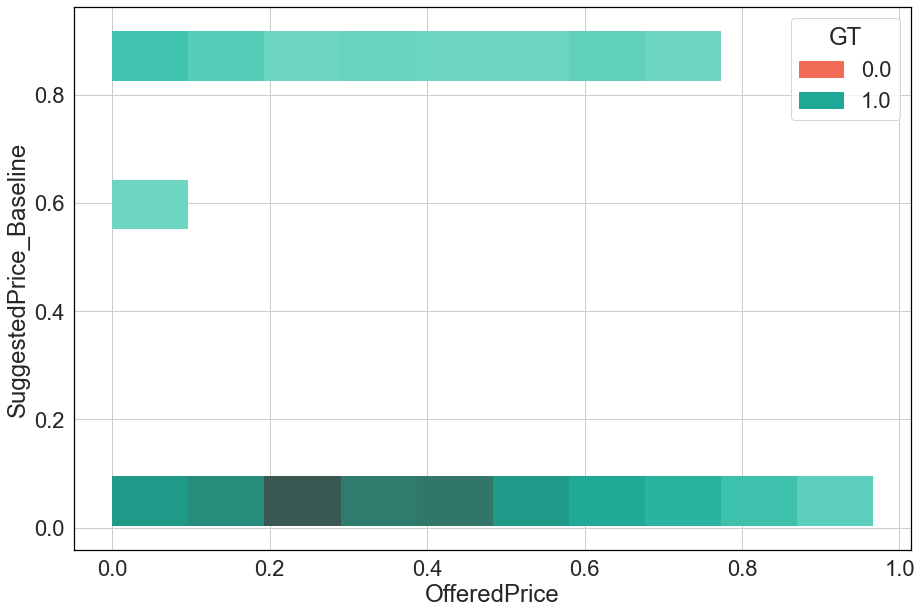} 
    \caption{Simulation baseline} 
    \label{heat:a} 
    \vspace{4ex}
  \end{subfigure}
  \begin{subfigure}[b]{0.5\linewidth}
    \centering
    \includegraphics[scale=0.2]{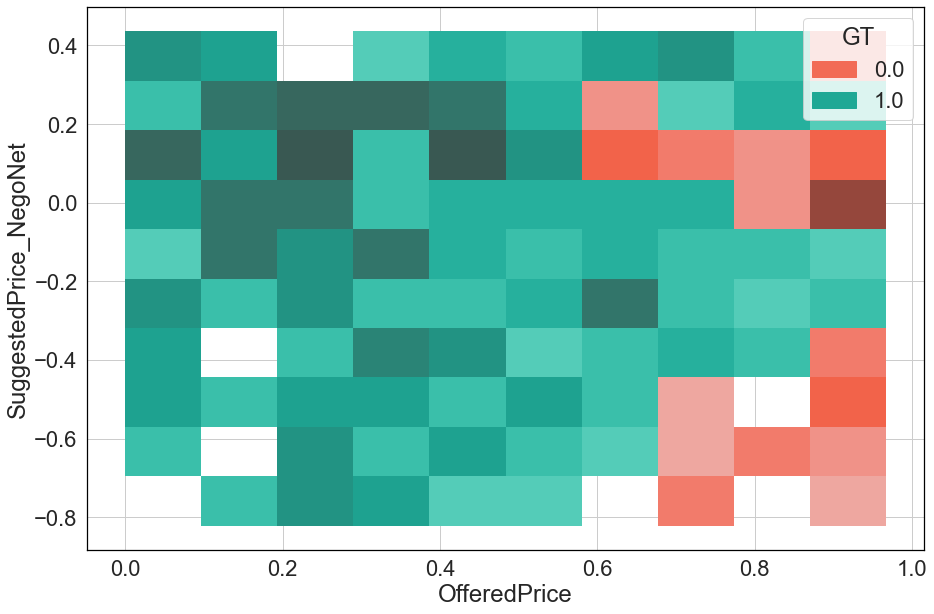} 
    \caption{Simulation NegoNets} 
    \label{heat:b} 
    \vspace{4ex}
  \end{subfigure} 
  \begin{subfigure}[b]{0.5\linewidth}
    \centering
    \includegraphics[scale=0.2]{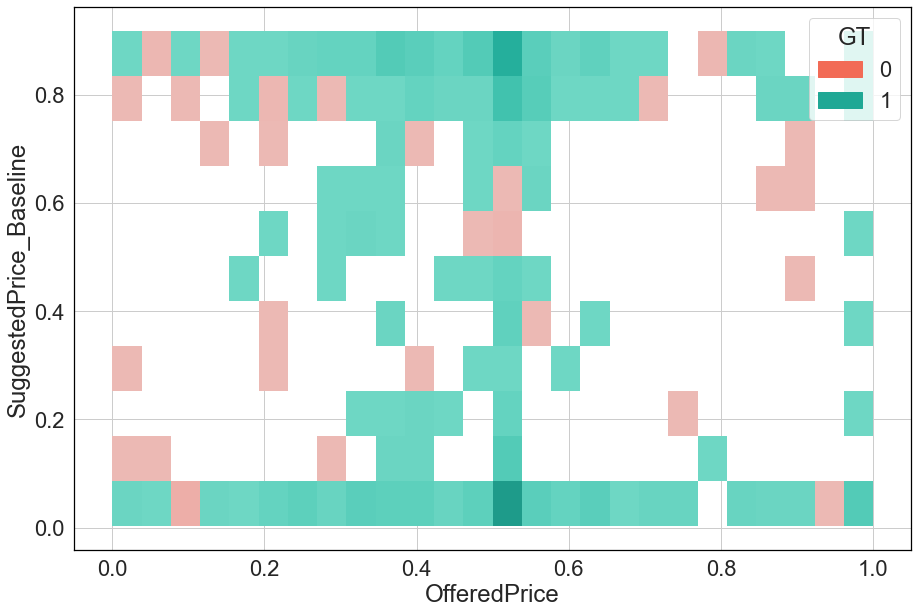} 
    \caption{Airline ancillary baseline} 
    \label{heat:c} 
  \end{subfigure}
  \begin{subfigure}[b]{0.5\linewidth}
    \centering
    \includegraphics[scale=0.2]{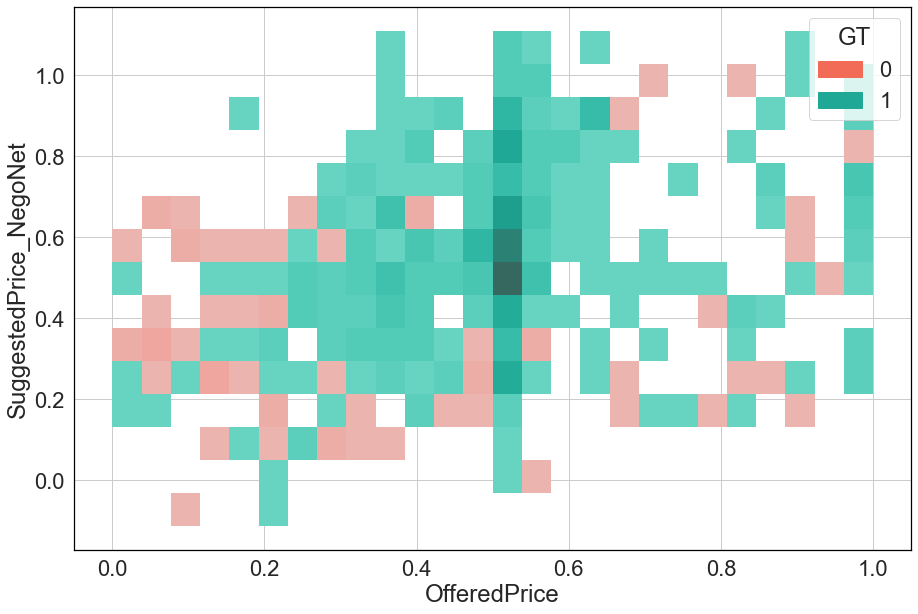} 
    \caption{Airline ancillary Negonets} 
    \label{heat:d} 
  \end{subfigure} 
  \caption{Heatmap of suggested prices with respect to offered prices (GT represent ground truth purchase indicator)}
  \label{heat} 
\end{figure}

To further investigate the qualitative performance of proposed framework, heatmaps of the suggested prices are shown in Figure \ref{heat}. For baseline simulation experiments, as shown in Figure \ref{heat:a}, the prices are clustered in case of baseline due to the classifier being confident in classification task. Also, NegoNets are suggesting heavy discounts including interactions where purchase is done (Figure \ref{heat:b}), which undesirable. Similarly, price distributions on airline ancillary dataset shown in Figure \ref{heat:c} and \ref{heat:d} implies that NegoNets tend to suggest variety of prices as compared to sequential classification based baseline method.
We plan to study the variance in offered prices influencing customer choice in future.

\end{document}